\documentclass[%
 reprint,
 amsmath,amssymb,
 aps,
]{revtex4-2}

\usepackage{graphicx}% Include figure files
\usepackage{dcolumn}% Align table columns on decimal point
\usepackage{bm}% bold math
\usepackage{booktabs}
\DeclareMathOperator{\tr}{tr}

\begin{document}

\title{Plastic Tensor Networks for Interpretable Generative Modeling}

\author{Katsuya O. Akamatsu}
%\email{akamatsu@issp.u-tokyo.ac.jp}
\affiliation{Institute for Solid State Physics, The University of Tokyo}
\author{Kenji Harada}
%\email{harada.kenji.8e@kyoto-u.ac.jp}
\affiliation{Graduate School of Informatics, Kyoto University}
\author{Tsuyoshi Okubo}
%\email{t-okubo@phys.s.u-tokyo.ac.jp}
\affiliation{Institute for Physics of Intelligence, The University of Tokyo}
\author{Naoki Kawashima}
%\email{kawashima@issp.u-tokyo.ac.jp}
\affiliation{Institute for Solid State Physics, The University of Tokyo}
\affiliation{Trans-scale Quantum Science Institute, The University of Tokyo}

\begin{abstract}
A structural optimization scheme for a single-layer nonnegative adaptive tensor tree (NATT) that models a target probability distribution is proposed as an alternative paradigm for generative modeling. The NATT scheme, by construction, automatically searches for a tree structure that best fits a given discrete dataset whose features serve as inputs, and has the advantage that it is interpretable as a probabilistic graphical model. We consider the NATT scheme and a recently proposed Born machine adaptive tensor tree (BMATT) optimization scheme and demonstrate their effectiveness on a variety of generative modeling tasks where the objective is to infer the hidden structure of a provided dataset. Our results show that in terms of minimizing the negative log-likelihood, the single-layer scheme has model performance comparable to the Born machine scheme, though not better. The tasks include deducing the structure of binary bitwise operations, learning the internal structure of random Bayesian networks given only visible sites, and a real-world example related to hierarchical clustering where a cladogram is constructed from mitochondrial DNA sequences. In doing so, we also show the importance of the choice of network topology and the versatility of a least-mutual information criterion in selecting a candidate structure for a tensor tree, as well as discuss aspects of these tensor tree generative models including their information content and interpretability.
\end{abstract}


\maketitle

\section{Introduction}
Tensor network (TN) methods are useful in the treatment of quantum many-body systems. Generally, TN approaches focus largely on Born-type ansatze where the TN represents some wavefunction $\Psi(\vec{x})$. Thus, the associated probability of some outcome $\vec{x}$ is obtained by taking the squared norm of the wavefunction according to the Born rule:
\begin{equation}
P(\vec{x})=\frac{|\Psi(\vec{x})|^2}{Z},\quad Z=\sum_{\{\vec{x}\}}|\Psi(\vec{x})|^2
\end{equation}
The problem of computing $|\Psi(\vec{x})|^2$, which is represented in TN notation by copying and reflecting the ansatz $\Psi(\vec{x})$, is dramatically simplified with the use of canonical forms available for ansatze like a matrix product state (also called a tensor train) or tensor tree, which are loop-free. Most TN machine learning approaches inherit this idea, and the use of TNs in generative modeling has been demonstrated with various types of ansatze and on a number of common benchmark datasets \cite{stoudenmire2016,han2018,cheng2019,ran2020,felser2021,wall2021}. However, while there is some work directly modeling the probability distribution, such as methods targeting 2D classical lattice spin models (like the tensor renormalization group algorithm \cite{levin2007}), methods for single-layer nonnegative MPS-type networks \cite{glasser2019} and methods for the construction of a tree approximation for Ising spin glass instances \cite{kawashima2006}, a more general data-driven tensor tree approach that does not rely on the Born rule has yet to be considered in the literature. In this work, we propose a single-layer nonnegative scheme for a tensor tree that directly models a given target distribution, as a complement to Born machine-based methods for tensor trees that require a double-layer architecture.

We stress that both approaches have merits: we found that the double-layer approach, which is quantum-inspired, tends to perform better overall when it comes to identifying a correlation structure, whereas a single-layer approach that models the probability directly using nonnegative tensors allows for the classical interpretation of classical datasets. Furthermore, it is known the Born machines on tensor trees are equivalent to quantum unitary circuits, and that nonnegative tensor trees are equivalent to hidden Markov models defined on a tree \cite{glasser2019}. This provides a guide as to where they might find a natural application as a model. While there are existing approaches based on tensor sketching for generative modeling purposes \cite{hur2023,peng2023}, they assume a fixed structure. There is also a sketching-based method that also adaptively determines the network topology of a graphical model given samples from the model \cite{tang2023}, but they only apply to cases where all nodes in the model are visible nodes. The class of Markov models that our proposed method targets contain hidden variables, and all input features are associated with terminal nodes.

The choice of structure in TN machine learning greatly impacts the outcome of the modeling task. Depending on the nature of the dataset, some structures can be more natural candidates compared to other schemes: one-dimensional data is modeled using a chain (an MPS) \cite{stoudenmire2016,han2018}, and two-dimensional image data can be represented better with a 2D tensor tree as opposed to a chain \cite{cheng2019}. Thus, one important aspect of generative modeling with TNs is how the geometry is to be selected: when the structure of the data is apparent, we can prescribe a good candidate. However, in general, given some data, it is hard to identify the hidden structure that best represents it. A recent method in this direction \cite{hikihara2023,harada2024} proposes to optimize the structure of a quantum state modeled with a tensor tree by selecting local connection geometries that minimize some information quantity defined for each bond in the network. We closely follow this principle in our proposed method to generate plastic structures that dynamically accommodate for input data and aim to demonstrate the versatility of this criterion and the general class of adaptive tensor tree (ATT) schemes on various types of data.

\begin{figure*}
\centering
\includegraphics[width=0.8\linewidth]{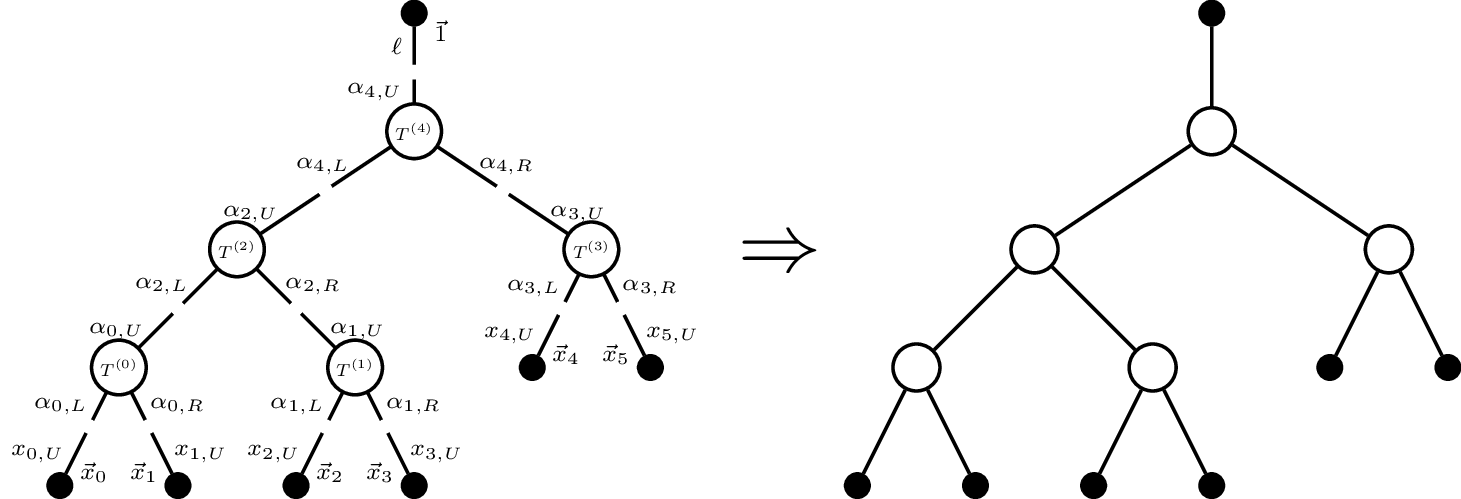}
\caption{Construction of a nonnegative tensor network representing the weight function $W(\vec{x})$. The tensor tree model is composed of nonnegative tensors $T^{(n)}$, and input vectors are labeled like $\vec{x}_i$. At the top bond, the vector of all ones $\vec{1}$ is supplied in the context of unsupervised modeling for a discrete data distribution where inputs are encoded as probabilities. For a Born machine, the tensor network elements are unconstrained and this construction represents the wavefunction $\Psi(\vec{x})$.}
\label{fig:tt_diagram}
\end{figure*}

We focus on binary tensor trees for modeling some discrete probability distribution $P$. A (full) binary tensor tree denoted as $\Lambda$ is comprised of three-legged tensors $\{T^{(n)}_{\alpha_{n,L},\alpha_{n,R},\alpha_{n,U}}\}\in\Lambda$, where $n$ is an index to track individual tensors, and $\alpha_{n,L},\alpha_{n,R},\alpha_{n,U}$ denote indices along the left, right and upper legs of the tensor indexed $n$, respectively. By correctly joining the legs of these tensors to form a network whose underlying structure is that of a tree graph, we obtain a graphical representation of the tensor tree $\Lambda$ (as depicted in Fig.~\ref{fig:tt_diagram}). If the binary tensor tree $\Lambda$ contains $N_I$ open legs for input vectors (and one additional open leg for a top vector composed of all ones in the context of unsupervised learning), then there are a total of $|\Lambda|=N_I-1$ internal degree-3 tensors in the tensor tree representation.

The weight function $W(\vec{x})$ can be expressed diagrammatically (Fig.~\ref{fig:tt_diagram}) by joining the open legs of the model $\Lambda$ with the legs of an input $\vec{x}$. The object $\vec{x}$ is an $N_I$-component vector of vectors that represents the input to the network. The associated probability distribution represented by the tensor tree $\Lambda$ can be obtained as:
\begin{equation}
P(\vec{x})=\frac{W(\vec{x})}{Z},\quad Z=\sum_{\{\vec{x}\}}W(\vec{x})
\end{equation}
Here, the sum in calculating the partition function $Z$ can be obtained by considering every possible valid input to $W(\vec{x})$ (and by extension, $\Lambda$). For a tensor tree, this can be done by contracting the vector whose elements are all ones with each input leg of the network.

We introduce a scheme based on a nonnegative matrix factorization (NMF) for the optimization of a single-layer nonnegative adaptive tensor tree (NATT) of arbitrary structure with the goal of modeling a target distribution described by an input dataset. We seek a tensor tree representation $\Lambda$ such that for all vectors in the space of possible inputs $\mathcal{X}$, the output weight $W$ is nonnegative:
\begin{equation}
W(\vec{x})\geq0\quad\forall\vec{x}\in\mathcal{X}
\end{equation}
While more generally, we could consider single-layer tensor tree models for $W(\vec{x})$ constrained so that they always produce a nonnegative output for any input, without directly imposing restrictions on each element, in practice, we are aware of two ways to impose this constraint: either we represent the whole network as a product of two copies of the identical network (which corresponds to a Born machine), or we must restrict each component tensor to have nonnegative elements. We describe a tensor tree as nonnegative when the elements of all its component tensors are nonnegative, so that we can guarantee that the output of the network is positive. This also has the added advantage that each tensor element is individually interpretable as a weight/probability: 
\begin{equation}
T^{(n)}_{\alpha_{n,L},\alpha_{n,R},\alpha_{n,U}}\geq0\quad\forall T^{(n)}\in\Lambda
\end{equation}
Note that explicitly computing the probability distribution from a Born machine representation can be done by considering two copies of the wavefunction and folding it on itself. This effectively squares the bond dimension. To represent the probability distribution $P$ in the Born machine approach, we must fold two copies of the wavefunction onto each other. If the wavefunction has bond dimension $\chi$, the resulting single-layer tensor tree will have bond dimension $\chi^2$.

In the wavefunction representation, for N tensors, the total number of parameters grows like $\mathcal{O}(\chi^3)$. For the folded single-layer tensor tree, the naive count of the total number of parameters is $\mathcal{O}(\chi^6)$ but they are all determined by $\mathcal{O}(\chi^3)$ parameters. However, there is no guarantee that the resulting single-layer network contains exclusively nonnegative tensors. This makes it impractical to use a Born machine representation when searching for a probabilistic explanation for a given dataset.

\section{NLL optimization of a NATT}
The proposed method is based on a recent method that accomplishes the structural optimization of a double-layer Born machine adaptive tensor tree \cite{harada2024} (which we abbreviate as BMATT), where the optimization is done by first contracting tensors along a bond, then optimizing the fused tensor. Various reconnection geometries are proposed by considering the three different ways to split the fused four-legged tensor into a pair of three-legged tensors, and the geometry that minimizes the mutual information, effectively decorrelating the subsystems, is selected. Our proposed NATT scheme follows the same type of criterion.

We intend to construct a generative model (a probability distribution $P$) represented as a single-layer nonnegative tensor tree. As in \cite{harada2024}, the cost function we use is the negative log-likelihood. For a given input dataset $X$, the negative log-likelihood $\mathcal{L}$ is:
\begin{equation}
\mathcal{L}=-\frac{1}{|X|}\sum_{\vec{x}\in X}\ln{P(\vec{x})}
\end{equation}

\begin{figure*}
\centering
\includegraphics[width=0.8\linewidth]{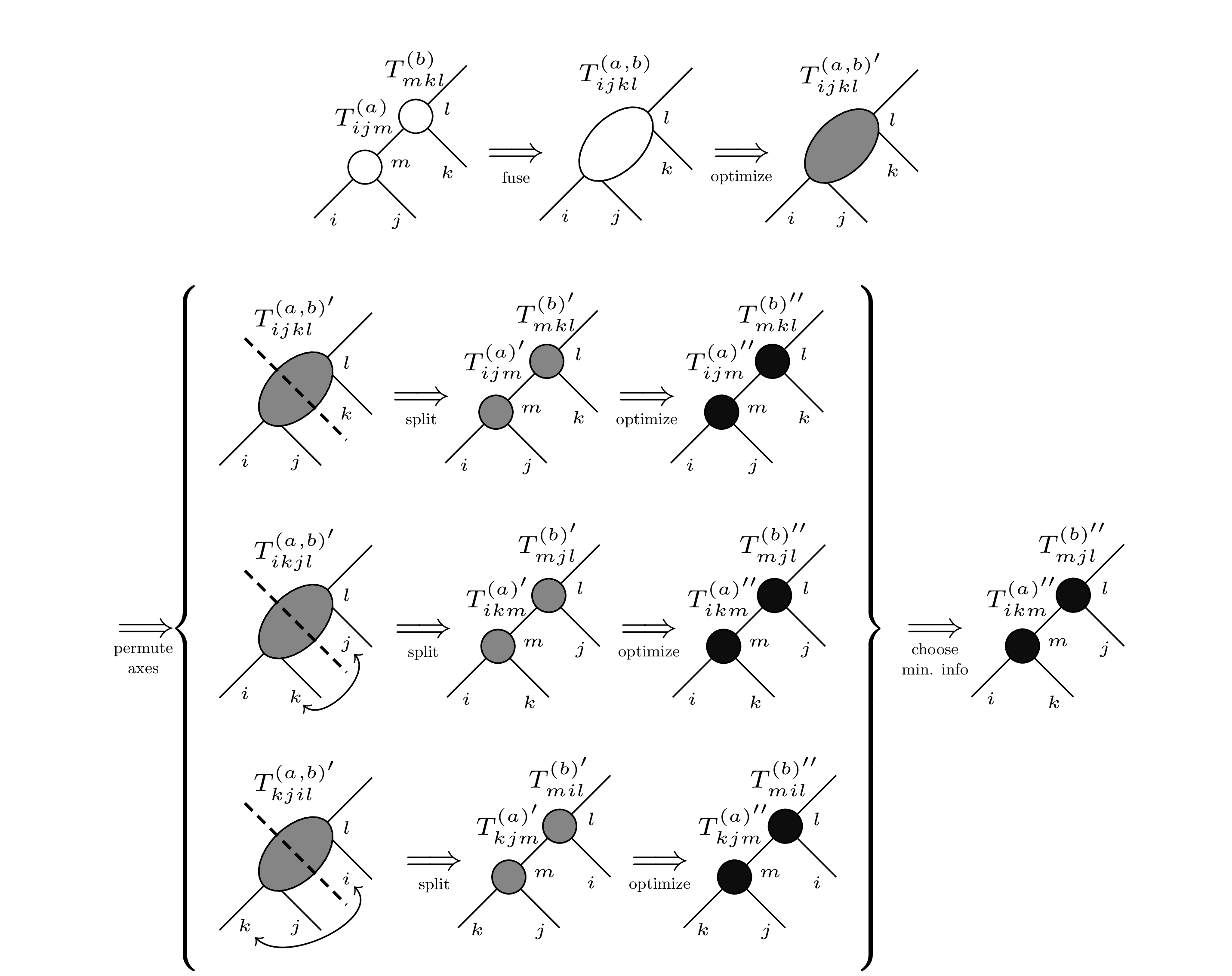}
\caption{Common structure of the ATT optimization schemes discussed in this work. In the proposed method, the optimization is done using projected gradient descent, the splitting is done using a nonnegative matrix factorization, and the connection geometry selection criterion minimizes the mutual information.}
\label{fig:alg}
\end{figure*}

A general framework for an ATT optimization scheme as laid out in previous work \cite{hikihara2023,harada2024} is described in Fig.~\ref{fig:alg}. The next subsections describe individual components of the proposed single-layer nonnegative scheme.

\subsection{Optimization of the fused tensor}
We consider two tensors $T^{(a)}_{ijm}$ and $T^{(b)}_{mkl}$, where $T^{(a)}$ is the downstream tensor (the topmost tensor serves as the root of the tree), linked by a bond somewhere on the tree. The tensors are contracted to obtain a four-legged fused tensor which we denote as $T^{(a,b)}_{ijkl}$ (here, we assume without loss of generality that the common bond is the left leg of the upstream tensor and the upper leg of the downstream tensor):
\begin{equation}
T^{(a,b)}_{ijkl}=\sum_mT^{(a)}_{ijm}T^{(b)}_{mkl}
\end{equation}
Unlike in the case of the Born machine, this fused tensor must be optimized while preserving the nonnegativity of its elements. We apply a form of projected gradient descent using the nonnegative projection operator $\bm{P}_{NN}(x)\equiv\max(0,x)$:
\begin{gather}
g_{\bm{P}}=T^{(a,b)}_{ijkl}-\bm{P}_{NN}(T^{(a,b)}_{ijkl}-\nabla_{T^{(a,b)}_{ijkl}}\mathcal{L}) \nonumber \\
=\begin{cases}\nabla_{T^{(a,b)}_{ijkl}}\mathcal{L},\quad T^{(a,b)}_{ijkl}>\nabla_{T^{(a,b)}_{ijkl}}\mathcal{L} \\ T^{(a,b)}_{ijkl},\quad T^{(a,b)}_{ijkl}\leq \nabla_{T^{(a,b)}_{ijkl}}\mathcal{L} \end{cases} \\
T^{(a,b)'}_{ijkl}=\bm{P}_{NN}(T^{(a,b)}_{ijkl}-\eta g_{\bm{P}})
\end{gather}
$\eta$ is the learning rate, $\nabla_{T^{(a,b)}_{ijkl}}\mathcal{L}$ is the gradient with respect to the tensor element, and $g_{\bm{P}}$ is the projected gradient.
For the single-layer tensor network $\Lambda$ and with respect to the fused tensor $T^{(a,b)}_{ijkl}$, the gradient is:
\begin{equation}
\nabla_{T^{(a,b)}_{ijkl}}\mathcal{L}=\frac{1}{Z}\nabla_{T^{(a,b)}_{ijkl}}Z-\frac{1}{|X|}\sum_{\vec{x}\in X}\frac{1}{W(\vec{x})}\nabla_{T^{(a,b)}_{ijkl}}W(\vec{x})
\end{equation}
The derivatives $\nabla_{T^{(a,b)}_{ijkl}}Z$ and $\nabla_{T^{(a,b)}_{ijkl}}W(\vec{x})$ can be obtained by excluding the fused tensor from the diagrams for $Z$ and $W(\vec{x})$ respectively and contracting the network.
In our implementation, we combine the projected gradient scheme with gradient clipping by norm (clipping to unit Frobenius norm) and use the AdamW optimizer \cite{loshchilov2017}.

\subsection{NMF and optimization of individual tensors}
For the Born machine, the SVD is used to recover two three-legged tensors from the fused four-legged tensor. However, the SVD does not respect the nonnegativity constraint. Instead, we compute a nonnegative matrix factorization (NMF) of the optimized fused tensor $T^{(a,b)'}_{ijkl}$. For now, we assume that no changes are made to the structure. That is, we find two nonnegative tensors $T^{(a)'}_{ijm}$ and $T^{(b)'}_{mkl}$ so that:
\begin{equation}
T^{(a,b)'}_{ijkl}\approx\sum_mT^{(a)'}_{ijm}T^{(b)'}_{mkl}
\end{equation}
Denoting the dimension of an index as $\chi$, the dimension of the index $m$ is upper-bounded by the minimum dimensions of the other two pairs of indices taken jointly: $(ij),(kl)$ with dimensions $\chi_i\chi_j,\chi_k\chi_l$ respectively, so that $\chi_m=\min(\chi_i\chi_j,\chi_k\chi_l)$. In practice, we must enforce an upper bound on the bond dimension $\chi_{\text{max}}$, so we compute the NMF with $\chi_m=\min(\chi_i\chi_j,\chi_k\chi_l,\chi_{max})$. 

The NMF is computed by applying multiplicative updates to the factor matrices, minimizing the KL divergence as defined in \cite{lee2000} between the target matricized fused tensor and its reconstruction. For the NMF approximation $V\approx WH$, the KL divergence is:
\begin{equation}\label{eq:mat_kl}
\mathcal{L}_{mat}=\sum_{ij}\left((WH)_{ij}-V_{ij}+V_{ij}\log\frac{V_{ij}}{(WH)_{ij}}\right)
\end{equation}
We considered other NMF schemes minimizing the matrix KL divergence \cite{hsieh2011,hien2021}, but we found that the multiplicative update (MU) scheme worked best. The initial condition for the NMF computation was based on the SVD of the target matrix following the prescription in \cite{boutsidis2008}.

The computational costs of the BMATT and NATT schemes are dominated by the cost of the SVD and NMF respectively. Since the NMF procedure in our implementation is initialized using an SVD-based approach, the NATT scheme is more expensive than the BMATT scheme. Furthermore, because the NMF is iteratively computed, the cost of the NATT approach is practically determined by the convergence speed of the iterative method. We found that the number of MU iterations needed for convergence grows as size of the matrix increases. When the maximum bond dimension is $\chi_{max}=2$, the NATT scheme is almost as fast as the BMATT scheme, with the gap between them increasing as $\chi_{max}$ increases. In terms of $\chi_{max}$, the cost of the truncated SVD keeping $\chi_{max}$ singular values scales like $\mathcal{O}(\chi_{max}^5)$ \cite{halko2011} (assuming a fused tensor whose four legs have bond dimension $\chi_{max}$). However, in our implementation, we computed the full SVD with cost $\mathcal(\chi_{max}^6)$. MU iterations minimizing the KL divergence for computing the NMF of the matricized fused tensor have cost $\mathcal{O}(\chi_{max}^5)$ per iteration \cite{hsieh2011}.

For the NATT scheme, the bottleneck depends on the number of iterations used to compute the NMF. MU tends to converge slowly for dense and unstructured matrices, and empirically, we found that for randomly-generated pattern learning problems, the ratio between the time needed to run the NATT and BMATT schemes can grow rapidly: when $\chi_{max}=2$, the ratio is typically close to 1, but when $\chi_{max}=10$, the NATT can be a hundred times more expensive (we observed a maximum of around 127). For random data at $\chi_{max}=2$, the ratio ranged from 1.8 to 2.5. For random data at $\chi_{max}=4$, the ratio ranged from 2.4 to 6. On the other hand, for structured data with $\chi_{max}=2$, we found that the ratio was typically close to 1 (we observed a maximum of around 1.7), and for a real-world dataset with $\chi_{max}=4$, the ratio was always less than 2 (we observed ratios between 1.1 to 1.6).

After splitting the fused tensor into three-legged tensors, we then apply the same nonnegative gradient descent strategy. For a three-legged tensor $T^{(a)}$, the following update is performed:
\begin{gather}
g_{\bm{P}}=T^{(a)'}_{ijm}-\bm{P}_{NN}(T^{(a)'}_{ijm}-\nabla_{T^{(a)'}_{ijm}}\mathcal{L}) \nonumber \\
=\begin{cases}\nabla_{T^{(a)'}_{ijm}}\mathcal{L},\quad T^{(a)'}_{ijm}>\nabla_{T^{(a)'}_{ijm}}\mathcal{L} \\ T^{(a)'}_{ijm},\quad T^{(a)'}_{ijm}\leq \nabla_{T^{(a)'}_{ijm}}\mathcal{L} \end{cases} \\
T^{(a)''}_{ijm}=\bm{P}_{NN}(T^{(a)'}_{ijm}-\eta g_{\bm{P}})
\end{gather}
The gradient is computed similarly to the case for the fused tensor:
\begin{equation}
\nabla_{T^{(a)'}_{ijm}}\mathcal{L}=\frac{1}{Z}\nabla_{T^{(a)'}_{ijm}}Z-\frac{1}{|X|}\sum_{\vec{x}\in X}\frac{1}{W(\vec{x})}\nabla_{T^{(a)'}_{ijm}}W(\vec{x})
\end{equation}
Again, the derivatives $\nabla_{T^{(a)'}_{ijm}}Z$ and $\nabla_{T^{(a)'}_{ijm}}W(\vec{x})$ can be obtained by excluding the tensor from the relevant diagrams. We optimize the two tensors in this manner ten times each, in an alternating fashion. As in the case for the fused tensor, we applied both gradient clipping by norm and the AdamW optimizer \cite{loshchilov2017} with the projected gradient descent step.

\subsection{Structural optimization based on mutual information}
While the basic NATT optimization scheme can already be described with the previous steps, we are also interested in determining a good model structure. Like in \cite{harada2024}, to optimize the structure of the network, we attempt to minimize the mutual information across all bonds. This is done by considering the three different ways to split a four-legged fused tensor, performing the gradient descent updates on each, and then evaluating the bond mutual information at the central bond linking the two considered tensors.

The mutual information (MI) between subsystems $A$ and $B$ is defined as a sum of entropies:
\begin{gather}
I(A,B)\equiv\sum_{\vec{a}\in\mathcal{X}_A}\sum_{\vec{b}\in\mathcal{X}_B}P(\vec{a},\vec{b})\log\frac{P(\vec{a},\vec{b})}{P(\vec{a})P(\vec{b})} \nonumber \\
=H_A+H_B-H_{AB}
\end{gather}
$\mathcal{X}_A$ and $\mathcal{X}_B$ are the sets of all possible inputs for subsystems $A$ and $B$ respectively. $H_{AB}$ is the Shannon entropy of the joint distribution of both $A$ and $B$, while $H_A$ and $H_B$ are obtained by tracing out the other subsystem and computing the marginal entropy. As the number of terms to consider in the sum grows exponentially, we estimate these quantities using the input dataset $X$. By using the input dataset to compute the estimated entropies $\tilde{H}_{AB},\tilde{H}_{A},\tilde{H}_{B}$, we avoid the exponential cost as well as spurious information that arises when the network is far from correct (such as in the initial stages of the optimization). We split the components of $\vec{x}\in X$ by the subsystem they belong to, like $\vec{x}=(\vec{x}_A,\vec{x}_B)$, and we write the vector of ones (really a vector of vectors of ones, since each input site takes a vector) as $\vec{1}$.
\begin{gather}
\tilde{H}_{AB}=-\frac{1}{|X|}\sum_{\vec{x}\in X}\ln{P_{AB}(\vec{x})} \\
\tilde{H}_A=-\frac{1}{|X|}\sum_{\vec{x}\in X}\ln{P_A(\vec{x})}=-\frac{1}{|X|}\sum_{\vec{x}\in X}\ln{P_{AB}((\vec{x}_A,\vec{1}_B))} \\
\tilde{H}_B=-\frac{1}{|X|}\sum_{\vec{x}\in X}\ln{P_B(\vec{x})}=-\frac{1}{|X|}\sum_{\vec{x}\in X}\ln{P_{AB}((\vec{1}_A,\vec{x}_B))}
\end{gather}
These quantities can be computed efficiently because the structure of the network is that of a tree. At this point, all references to the empirical MI refer to the MI estimated against the input dataset. After estimating the MI associated with each geometry, the connection geometry with the lowest bond mutual information is selected to replace the fused tensor.

In the case of the Born machine, we have access to the entanglement entropy (EE) $S_{EE}$ in addition to the MI. On a tensor tree, the EE at a bond can be calculated by summing over the squared Schmidt values of the bipartition $(A,B)$ associated with the bond. When using a Born machine representation, we are modeling a pure quantum state, so that:
\begin{gather}
S_{EE}(\rho_{AB})\equiv-\tr[\rho_A\log{\rho_A}]=-\tr[\rho_B\log{\rho_B}] \nonumber \\
=-\sum_i a_i^2\log a_i^2
\end{gather}
$\rho_A$ and $\rho_B$ are reduced density matrices tracing out degrees of freedom in $B$ and $A$ respectively. The variables $\{a_i\}$ correspond to the Schmidt singular values obtained by an SVD of the fused tensor associated with a bond (note that we are considering only one of the two layers). There is a relationship between the MI ($I(A,B)$), EE ($S_{EE}(\rho_{AB})$), and the bond dimension $\chi$ at a bond \cite{convy2022,wu2009}:
\begin{equation}
0\leq I(A,B)\leq S_{EE}(\rho_{AB})\leq\log\chi
\end{equation}
To quantify the information content of the model, we use the total bond MI $I_\Sigma$ and total bond EE $S_{EE,\Sigma}$ summed across all internal bonds in the tensor tree.

For the NATT scheme, there is no notion of entanglement entropy since the framework is entirely classical. Thus, we have the following inequality:
\begin{equation}
0\leq I(A,B)\leq\log\chi
\end{equation}
The first portion of the inequality comes from the nonnegativity of the MI and the second portion of the inequality is a consequence of the data processing inequality: by defining an intermediate variable $C$ passed on the bond between subsystems $A$ and $B$ and assuming a Markov property (variables are only influenced by connected variables on the tree), we have $I(A,B)\leq I(A,C)$. Since we have $I(A,C)\leq\log\chi$, we must have $I(A,B)\leq\log\chi$.

\subsection{Information content and interpretability}
One major difference between the NATT and BMATT models concerns their interpretability. Since the NATT is constrained to be nonnegative elementwise, we can directly interpret the elements as probability weights. In fact, to obtain the equivalent hidden Markov model that corresponds to the NATT, we can apply a nonnegative CP decomposition (NNCPD) on each tensor in the network, which can be done efficiently as the problem can be recast as multiple NMF subproblems for each factor matrix \cite{kolda2009}. The NNCPD can be computed using methods generalizing the MU iteration \cite{shashua2005} or using a hierarchical least squares approach \cite{gillis2012}. Then the delta tensors correspond to nodes in the hidden Markov model, and the matrices correspond to the transition matrices in the hidden Markov model.

However, for the BMATT, we can have negative elements in the tensors, and this interpretation is not directly possible. If a BMATT and NATT model the same distribution with the same network structure, the total bond MI associated with both networks should coincide. Since we are considering the classical probability distribution function as the target, there must be a non-negative tensor tree that exactly represents it (provided, of course, that there is no limit in the bond dimension). Assuming that the BMATT representation has a total bond EE that coincides with the total bond MI, we conjecture that there exists a sequence of local unitary operations applied along existing edges that can be performed on the bonds of the BMATT that would transform the BMATT into an NATT, but how this can be accomplished remains an open question. In any case, it seems that bringing a BMATT into a form where it can be readily interpreted classically does not appear to be a trivial task.

\section{Results}
We evaluated the performance of three types of schemes: the BMATT scheme proposed in \cite{harada2024}, the NATT scheme described earlier, and a hybrid scheme where the network is first trained as an BMATT, then the resulting initial network structure is used as an initial guess for a second training stage using the NATT scheme. In this hybrid scheme, the elements of the tensor tree after the first stage of training are reinitialized to enforce nonnegativity.

\begin{table*}
\caption{Training parameters for datasets used in this work. $N_{instances}$ is the number of problem instances (target distributions), $N_{trials}$ is the number of trials per instance, $N_{samples}$ is the number of samples per problem instance, and $N_{batch}$ is the batch size used in training. $\eta$ is the base learning rate, $t_{max}$ is the maximum number of iterations, $\chi_{input}$ is the bond dimension at input sites, $\chi_{max}$ is the maximum bond dimension of the network, and $L$ is the number of input sites.}
\label{tab:train_params}
\centering
\begin{tabular}{c|ccccc}
\toprule
 & Random & LRCorr & Bitwise & BayesNet & mtDNA \\
\midrule
$N_{instances}$ & $10$ & $10$ & $1$ & $1$ & $1$ \\
$N_{trials}$ & $1$ & $1$ & $10$ & $10$ & $10$ \\
$N_{samples}\equiv|X|$ & $10$ & $10$ & $1000$ & $10^6$ & $1140$ \\
$N_{batch}$ & $10$ & $10$ & $1000$ & $10^4$ & $1140$ \\
$\eta$ & $0.005$ & $0.005$ & $0.05$ & $0.005$ & $0.0005$ \\
$t_{max}$ & $10^4$ & $10^4$ & $10^4$ & $10^4$ & $10^4$ \\
$\chi_{input}$ & $2$ & $2$ & $2$ & $2$ & $4$ \\
$\chi_{max}$ & $2,4,6,8,10$ & $10$ & $2$ & $2$ & $4$ \\
$L$ & $64$ & $64$ & $48$ & $16$ & $16$ \\
\bottomrule
\end{tabular}
\end{table*}

We present the experiments in an order that reflects the increasing complexity of the underlying hidden structure of the problem. We consider the problems of learning random patterns (no structure), resolving the structure of data with long-range correlations (some correlation structure), modeling deterministic bitwise operations (a shallow structure), proposing likely explanations for the hidden internal structure of a random probabilistic Bayesian network (a hidden internal structure), and a real-world example in phylogenetics where we model hierarchically-clustered genetic sequence data (a hidden internal hierarchical structure). The parameters for each numerical experiment are listed in Tab.~\ref{tab:train_params}.

\subsection{Random data}
We considered the problem of learning random data as an initial benchmark to demonstrate that the methods we are examining perform as intended and minimize the loss function. We generated synthetic datasets with $|X|=10$ random $L=64$-bit strings and tested the performance of the training for the maximum rank $\chi_{max}\in\{2,4,6,8,10\}$. At $\chi_{max}=10$, the maximum bond dimension equals the number of samples, and it is expected that any network structure can perfectly represent the dataset. For this example, we averaged over 10 instances, with each instance using an independently-generated dataset. The maximum number of iterations was $n=10000$, and the base learning rate was set to $\eta=0.005$.
\begin{figure}
\centering
\includegraphics[width=0.99\linewidth]{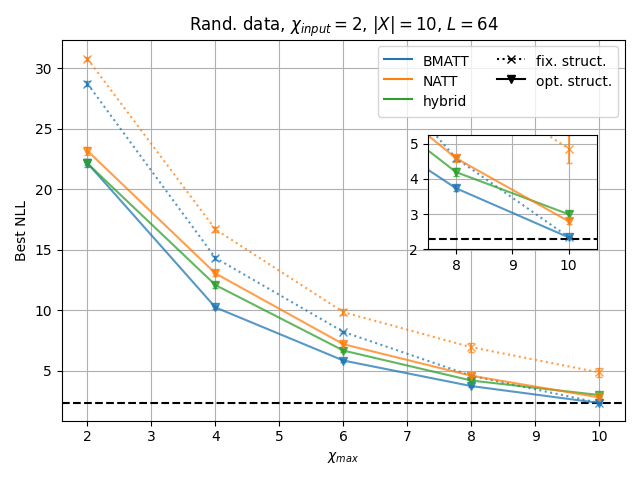}
\caption{Plot of the average NLL obtained over 10 instances when training a tensor tree on 10 samples of 64 random bits. Symbols denote whether or not the structure was also optimized in the process, the colors denote the training scheme. The horizontal dotted line represents the lower bound for the NLL, which is the NLL of the dataset.}
\label{fig:rand_l64}
\end{figure}

In Fig.~\ref{fig:rand_l64}, we plot the NLL as a function of the maximum bond dimension. For all three schemes, the NLL decreases as a function of rank, which demonstrates that the scheme works and minimizes the NLL as intended. Note that the hybrid scheme without structural optimization is omitted because it is essentially the same as the NATT without structural optimization. For the BMATT scheme, at $\chi_{max}=10$, the NLL approaches the bound. Without structural optimization, the NATT scheme does not consistently reach the global minimum even at $\chi_{max}=10$ since the optimization problem is generally harder due to the constraint and suboptimal structure. However, by optimizing the structure of the network, the NATT and hybrid schemes also achieve significantly lower NLL values and approach the bound at $\chi_{max}=10$. This demonstrates that applying the least-MI principle to choose a structure greatly improves the performance of this class of tensor tree-based ML methods, and that a proper choice of structure is desirable and impacts the achievable NLL.

\begin{figure}
\centering
\includegraphics[width=0.99\linewidth]{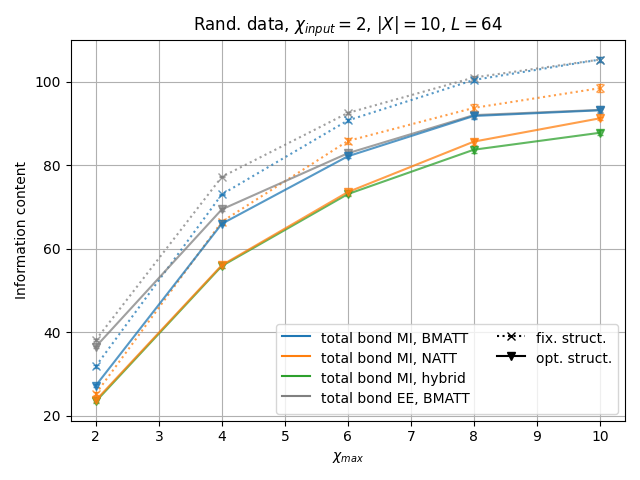}
\caption{Plot of the information content (summed over bonds) obtained over 10 instances, on 10 samples of 64 random bits. Note that all models admit a mutual information obtained from the modeled probability distribution, but the entanglement entropy (in gray) is also defined for the BMATT.}
\label{fig:rand_l64_info}
\end{figure}
Fig.~\ref{fig:rand_l64_info} contains a plot of the information content for each scheme. For the NATT and hybrid schemes, only the total bond MI $I_\Sigma$ is shown, and they are consistent with each other. In the case of the BMATT, both $I_\Sigma$ and $S_{EE,\Sigma}$ are plotted, and while there is initially a gap between $I_\Sigma$ and $S_{EE,\Sigma}$ at low rank, at $\chi_{max}=10$, we observe that both measures coincide, and that the estimated information content is similar to that of the NMF-based approaches. Note that when $\chi_{max}=10=|X|$, we can explicitly construct a nonnegative Born machine representation of the dataset by considering a tensor structure where each tensor index corresponds to one input data point. In a later example, we will consider a situation where $I_\Sigma$ and $S_{EE,\Sigma}$ do not coincide, even at the optimal rank for the dataset.

\subsection{Random data with long-range correlations}
In this problem, we considered small synthetic random datasets for which the middle bits are always fixed to either all-0s or all-1s and the left and right portions of the input data are randomly generated bits. We generated synthetic datasets with $|X|=10$ random $L=64$-bit strings, but this time the middle 32 bits are all either 0 or 1 (this means that the middle bits are always identical to each other). We tested the schemes with a $\chi_{max}=10$, so that there is a solution that saturates the NLL bound. The maximum number of iterations was $n=10000$, the base learning rate was set to $\eta=0.005$, and the initial condition for the tensor tree structure was set to be a random tree. This means that the tensor tree initially has no information on the structure of the data. The training parameters for this example are listed under the ``LRCorr'' column in Tab.~\ref{tab:train_params}. Since the sample size is small, each left substring is uniquely associated with a right substring and we have a dataset that exhibits a long-range correlation.

\begin{figure*}
\centering
\includegraphics[width=0.6\linewidth]{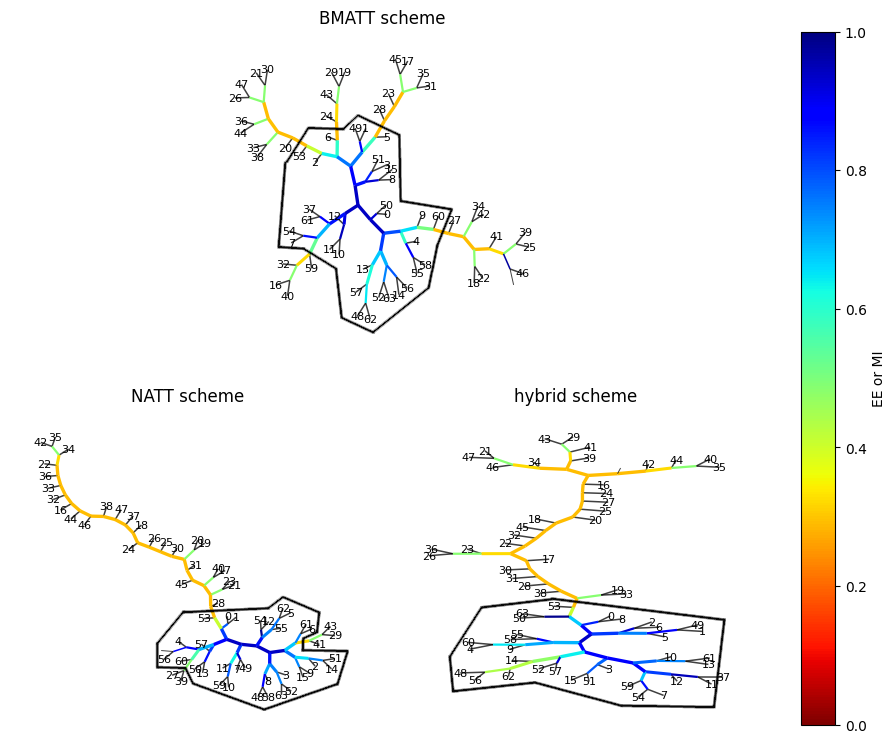}
\caption{Obtained network structures for a tensor tree trained on 10 samples of 64 random bits, where the middle 32 bits are always identical to each other and are either 0 or 1. Black lines separate the contiguous strongly-correlated portions on the left and right sides of the input string from the other inputs. For the BMATT, the bond colors denote the normalized EE ($S_{EE}/\ln\chi_{max}$), whereas for the NMF-based schemes, the bond colors correspond to the normalized MI ($I/\ln\chi_{max}$). In the network, positions 0-15 and 48-63 are the left and right subsections respectively, and sites 16-47 are the redundant middle section.}
\label{fig:rand_edges1d_l64_struct}
\end{figure*}
The objective for this example is for the network to be able to group together the strongly-correlated portion of the data. We show examples of obtained network structures in Fig.~\ref{fig:rand_edges1d_l64_struct}. For all schemes, a contiguous, strongly-correlated (as evidenced by the blue bond coloring indicating a large information content) portion of the network was obtained. This reproduces the result reported in \cite{harada2024} for the BMATT and also demonstrates that the same goal of clustering correlated portions closer to each other is achieved by the NATT and hybrid schemes.

\subsection{Binary bitwise operations}
Next, we considered synthetic datasets representing binary deterministic bitwise operations where the input is composed of $3L_{op}$ binary variables $(b_0,b_1,\cdots,b_{3L_{op}-1})$. They are constructed by generating mutually independent random binary numbers $b_0,b_1,\cdots,b_{2L_{op}-1}$ and then setting $b_{2L_{op}+i} := b_{L_{op}+i} \cdot b_{i}$ for $i=0,1,2,\cdots,L_{op}-1$, where $\cdot$ represents either the bitwise AND or the bitwise XOR operation. We generated datasets with $|X|=1000$ random $L_{op}=2,4,8,16$-bit strings. Here, we fixed $\chi_{max}=2$, which is sufficient to represent the target distribution, to see if the structure of the binary sentences could be captured by the schemes we are examining.

We consider two bitwise problems in this work, bitwise AND and bitwise XOR, but we stress that these methods are applicable to arbitrary operations (not necessarily with bits) that can be represented as a lookup table. From the earlier description, the true distribution should be described by $L_{op}$ clusters of size 3 (since the arity of the operation is 2, and we include the result bit). Each cluster should have elements whose indices are spaced by $L_{op}$ units. The maximum number of iterations was $n=10000$, the base learning rate was set to $\eta=0.05$, and the initial condition for the tensor tree structure was set to be a random tree.

The bitwise XOR problem is particularly unique in that there is a clear structure despite the observation that there are no two-point correlations between the three values $A,B,C$ in $A\oplus B=C$ for independently-drawn inputs $A$ and $B$. Thus, the problem is, in a sense, significantly harder than other binary bitwise operations like AND.
\begin{figure*}
\centering
\includegraphics[width=0.6\linewidth]{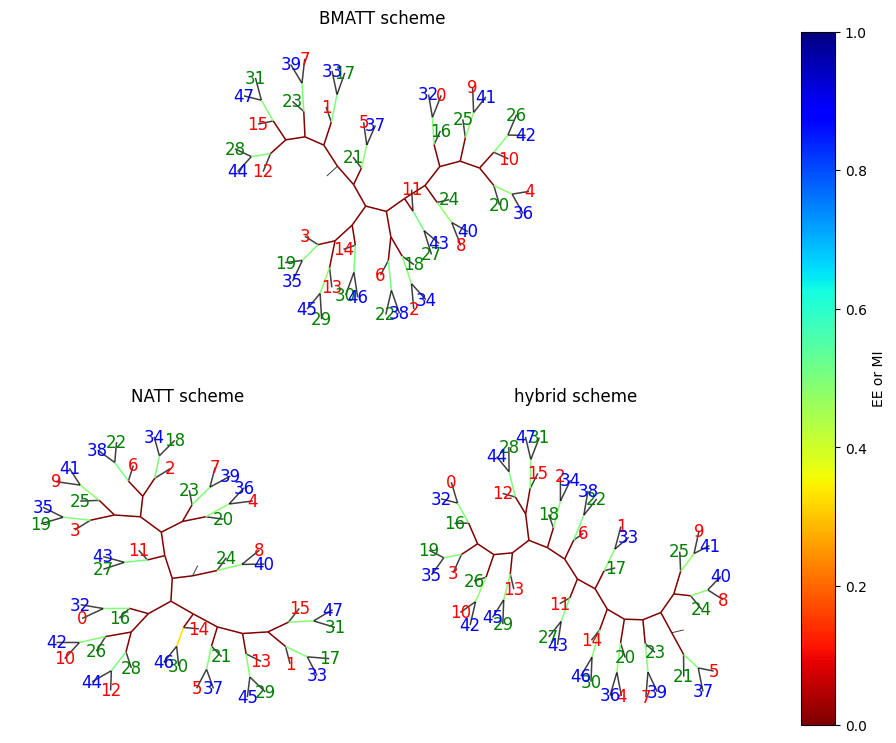}
\caption{Obtained network structures for a tensor tree trained on 1000 samples of 16-bit bitwise AND sentences (total of 48 input bits). The bond dimension is fixed to $\chi=\chi_{max}=2$. For the BMATT, the bond colors denote the normalized EE, whereas for the NMF-based schemes, the bond colors correspond to the normalized MI. Red and green labels denote bits in the first and second operands respectively, and blue labels denote bits in the result.}
\label{fig:and_train_i2_n1000_l16_rand_r2}
\end{figure*}

\begin{figure*}
\centering
\includegraphics[width=0.6\linewidth]{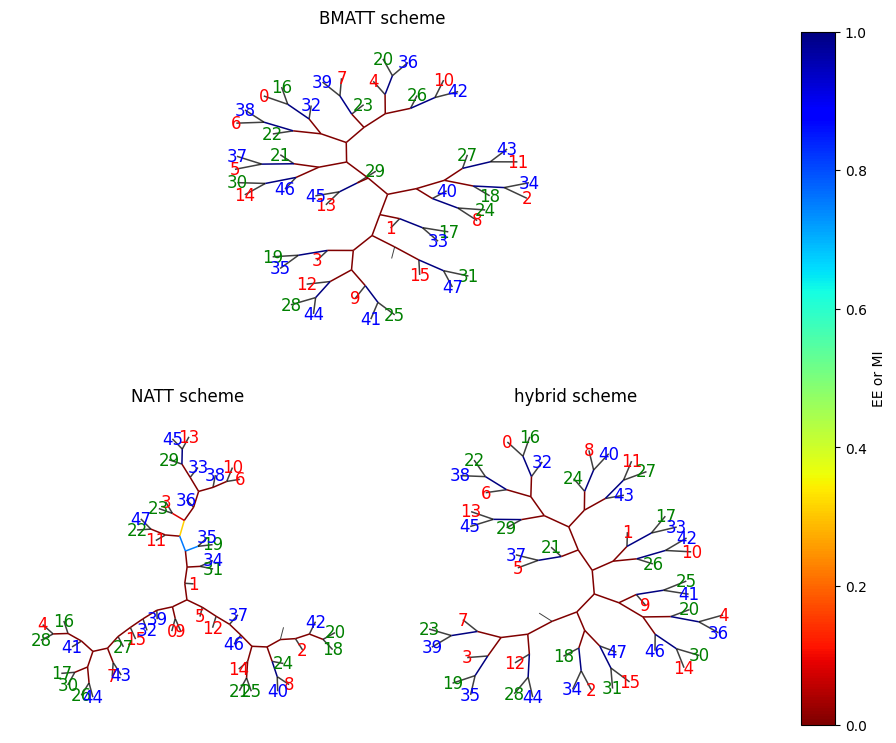}
\caption{Obtained network structures for a tensor tree trained on 1000 samples of 16-bit bitwise XOR sentences (total of 48 input bits). Only the BMATT and hybrid schemes provide correct structures. The bond dimension is fixed to $\chi=\chi_{max}=2$. For the BMATT, the bond colors denote the normalized EE, whereas for the NMF-based schemes, the bond colors correspond to the normalized MI. Red and green labels denote bits in the first and second operands respectively, and blue labels denote bits in the result.}
\label{fig:xor_train_i2_n1000_l16_rand_r2}
\end{figure*}
In Fig.~\ref{fig:and_train_i2_n1000_l16_rand_r2} and Fig.~\ref{fig:xor_train_i2_n1000_l16_rand_r2}, various obtained structures using the different tensor tree training schemes are shown. For the $L_{op}=16$ AND data, all of the methods are able to correctly cluster the input into 16 clusters, as all clusters, which are separated by bonds with zero MI/EE, consist of 3 leaf nodes whose positions differ by multiples of 16.

\begin{figure*}
\centering
\includegraphics[width=0.49\linewidth]{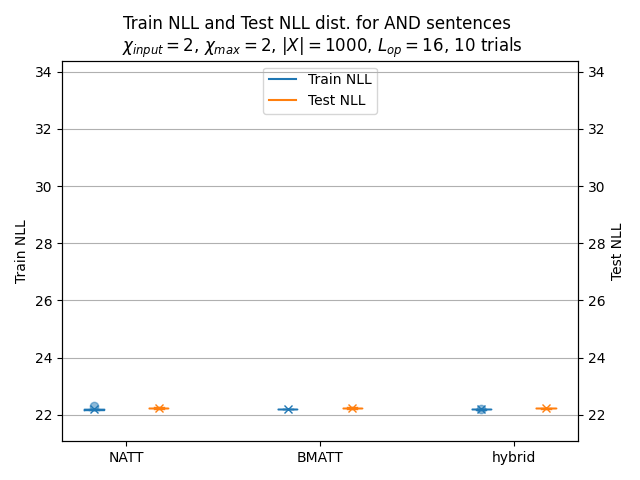}
\includegraphics[width=0.49\linewidth]{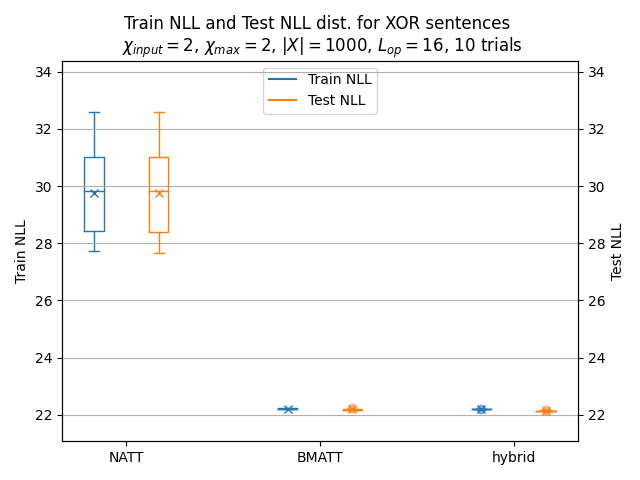}
\caption{Train and test NLL for 10 trials on AND and XOR data. The test dataset was a minimal set chosen to represent all possible valid sentences for each group of three bits. Circles indicate outliers, crosses indicate means, and the horizontal line in the boxes denote the median.}
\label{fig:bin_op_nll}
\end{figure*}

We observe the same result for the $L_{op}=16$ XOR data, except in the case of the standard NATT scheme, which has difficulty converging to the correct structure (Fig.~\ref{fig:bin_op_nll}). However, if the network is pretrained on the BMATT scheme and then trained using the NATT scheme, we find that the clustering remains correct. In addition to motivating the use of the hybrid scheme, this also suggests that the single-layer scheme, which models the probability distribution directly, may have difficulty finding a good choice of tree structure when there are little to no two-point correlations present in the target distribution. Intuitively, higher-order correlations are harder to detect, so in the absence of two-point correlations, the ideal structure would be harder to find.

Despite that, we find that when the nonnegative scheme is allowed to start from a good initial structure, it can find the correct clustering and avoid deviating from a correct structure. The tensors in the single-layer model linking correlated sites directly correspond to joint distributions over these sites that describe the target operation. These results suggest that the BMATT scheme is faster at finding a good correlation structure and justifies the utility of a hybrid scheme when taken together with the interpretability and classicality offered by the NATT scheme.

\begin{figure*}
\centering
\includegraphics[width=0.49\linewidth]{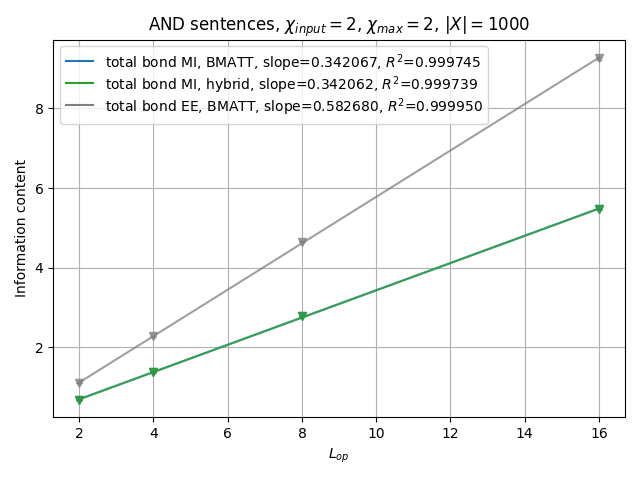}
\includegraphics[width=0.49\linewidth]{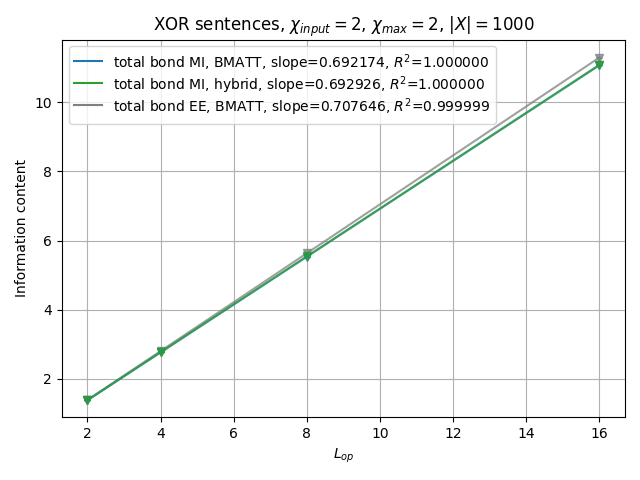}
\caption{Information content (summed over bonds) of the models trained on the bitwise AND (left) and XOR (right) synthetic datasets as a function of the input length.}
\label{fig:bin_op_info}
\end{figure*}
On a related note, we plot the information content of the models on both the AND and XOR data in Fig.~\ref{fig:bin_op_info}. Like the random data in Fig.~\ref{fig:rand_l64_info}, we see that for the XOR data, the total bond MI $I_\Sigma$ and total bond EE $S_{EE,\Sigma}$ are fairly close to each other. However, for the AND data, we observe that the $S_{EE,\Sigma}$ and $I_\Sigma$ are separated by a gap that widens as the operand length $L_{op}$ increases. To explain this, we note that for three bits $A$, $B$, and $C=A\land B$, there is a partitioning like $AB|C$ (here, $A$ and $B$ are on one side of a bond and $C$ is on the other side) where the total bond EE $S_{EE,\Sigma}$ and total bond MI $I_\Sigma$ match. However, in the partitionings $AC|B$ and $BC|A$, $I_\Sigma<S_{EE,\Sigma}$. Since we select the optimal network structure according to a least-MI principle, we observe this gap between $I_\Sigma$ and $S_{EE,\Sigma}$ for the AND data. In contrast, for the XOR data, all possible bipartitions yield the same EE and MI.

\begin{figure*}
\centering
\includegraphics[width=0.49\linewidth]{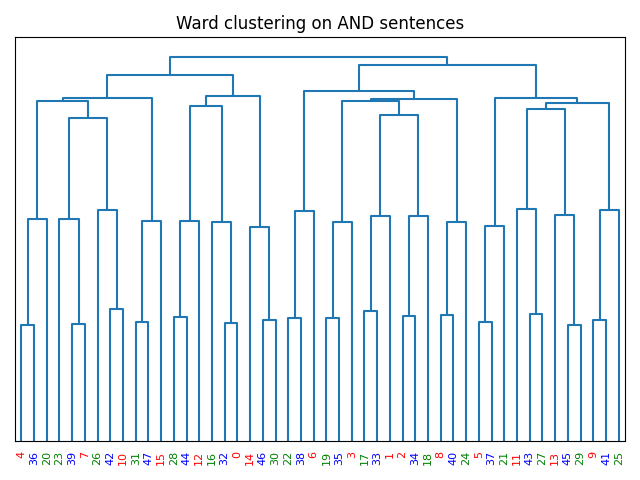}
\includegraphics[width=0.49\linewidth]{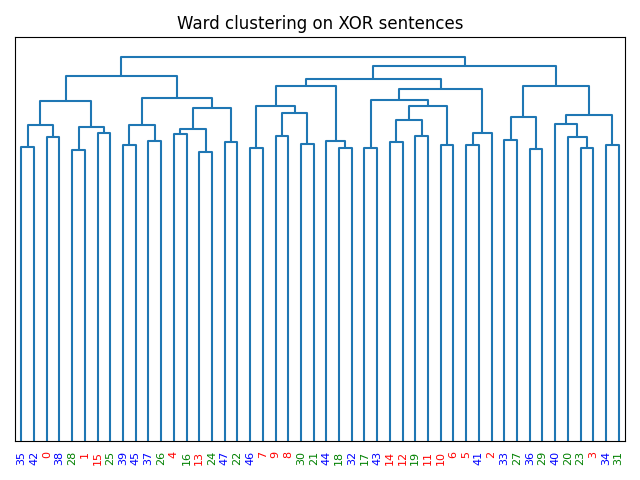}
\caption{Result of hierarchical clustering using the Ward method for the AND and XOR data. Distances were computed using the Hamming metric. The vertical axis is a measure of cluster distance. Note that the 16-bit AND data is clustered into input site triplets separated by 16, whereas there is no such order for the XOR data. Red and green labels denote bits in the first and second operands respectively, and blue labels denote bits in the result.}
\label{fig:ward_bin_op}
\end{figure*}
In Fig.~\ref{fig:ward_bin_op}, we show the result of Ward clustering \cite{ward1963}, which is a hierarchical clustering method minimizing the intracluster variance, with the input training data for the AND and XOR synthetic datasets. The results show that while the Ward clustering works for the AND data (since we obtain 16 clusters of size 3 whose indices are separated by 16), it fails for the XOR data. This is because the XOR data does not contain any two-point correlations, which means that approaches that rely on the calculation of distance matrices, which are essentially measures of two-point correlation, are bound to fail. In contrast, the ATT methods that we have considered (the BMATT and hybrid schemes) are able to treat the XOR correctly, indicating that these methods can capture higher-order correlations and work for pathological cases where there are no two-point correlations. Furthermore, the NATT and hybrid methods provide a means to deduce the operation by computing the joint distribution, even for hidden sites, whereas conventional hierarchical clustering methods do not provide this information.

\subsection{Random binary branching Bayesian networks}
\begin{figure}
\centering
\includegraphics[width=0.99\linewidth]{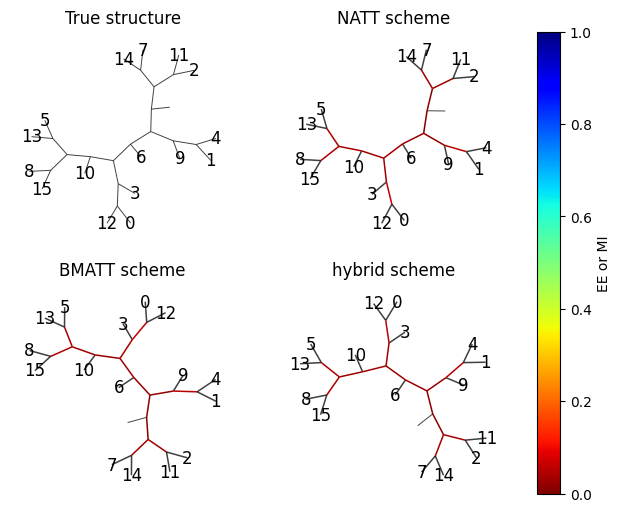}
\caption{Obtained network structures for a tensor tree trained on 100000 samples of a random Bayesian network defined on a full binary tree with 16 visible sites. For the BMATT, the bond colors denote the normalized EE, whereas for the NMF-based schemes, the bond colors correspond to the normalized MI. Positions 0-15 in the network are input sites.}
\label{fig:bayesian_rand16_n100000_struct}
\end{figure}

We then move on to random structured problems that can be divided into visible sites and hidden internal sites. Here, we consider a branching Bayesian network with 16 visible sites that is defined on a full binary tree (so that there are 15 hidden sites corresponding to internal tensors in the tensor tree). The internal structure of this model is randomly generated and then taken to be fixed (see Fig.~\ref{fig:bayesian_rand16_n100000_struct}, upper left). The dataset is composed of $|X|=100000$ samples obtained by randomly initializing the state of the top hidden site and maintaining it downstream with probability $p=0.8$. With this kind of internal Bayesian network topology, the aim is for the schemes to propose a model that describes the hidden structure of the data.

We then trained tensor tree models on it using the three schemes to either replicate the structure of the Bayesian model or obtain an equivalent model saturating the NLL of the dataset. The initial network structure is random and the maximum bond dimension was taken to be $\chi_{max}=2$, since the source model also has the same $\chi_{max}$. We set the maximum number of iterations to $N=10000$, used a batch size of 1000, and set the learning rate to $\eta=0.001$.

In Fig.~\ref{fig:bayesian_rand16_n100000_struct}, we show a typical result from each of the schemes and compare it to the true structure of the generating Bayesian network. All of the schemes typically produce structures that match or are very close to the true topology of the randomly-generated Bayesian network. It can also be seen that the site ordering is also generally respected, which suggests that the models have correctly learned the structure of the Bayesian network.

However, one could also measure the consistency of the method and how distant the generated structures are with each other for a given scheme. To accomplish this, we use a measure of tree distance grounded in information theoretic principles, the normalized cluster information distance (CID) \cite{smith2020} (see Appendix~\ref{app:cid}). A smaller value for the CID indicates that trees agree well with the hierarchical grouping of leaf nodes, and a larger value indicates disagreement between trees. A CID of 0 indicates perfect agreement between two tree graph topologies. Note that the value of the normalized CID grows fast initially as a function of the number of moves needed to bring the trees into agreement, but tapers off as the trees are more different from each other.

\begin{figure}
\centering
\includegraphics[width=0.99\linewidth]{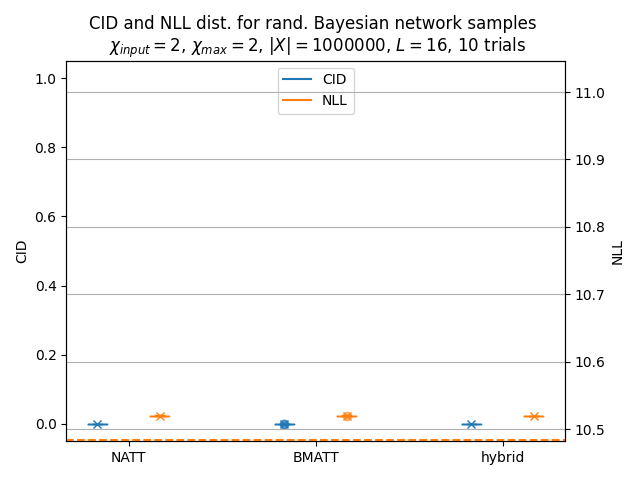}
\caption{Box plots of the cluster information distance (CID) (blue) and NLL (orange) for the models trained on a Bayesian random network. We considered one instance of the problem, and for each scheme, 10 trials were done. The NLL box plots consist of those 10 samples, while the CID box plots are taken across all pairs of trials, for a total of 45 (ways to choose 2 from 10) samples. Circles indicate outliers, crosses indicate means, and the horizontal line in the boxes denote the median. The dashed line represents the NLL of the dataset.}
\label{fig:att_bayesian_rand_tree_dist_cid_nll}
\end{figure}
In Fig.~\ref{fig:att_bayesian_rand_tree_dist_cid_nll}, we considered a single instance of a randomly-generated branching Bayesian network, and we plot the CID and NLL across 10 trials for each scheme. The CID was computed for each pair of distinct trials, for a total of 45 data points per box plot, and the NLL box plot uses the results of each of the 10 trials. All schemes generally achieve good NLLs that are close to each other. Since the obtained CIDs were all zero, all trials produced the same tree structure and we conclude that the methods are all capable of successfully reproducing the target structure given only a subset of the states in the Bayesian network, in a consistent fashion. We again note that the NMF-based methods produce models that can be directly interpreted as a probability distribution, which is a task that may be nontrivial in the BMATT representation.

\begin{figure}
\centering
\includegraphics[width=0.99\linewidth]{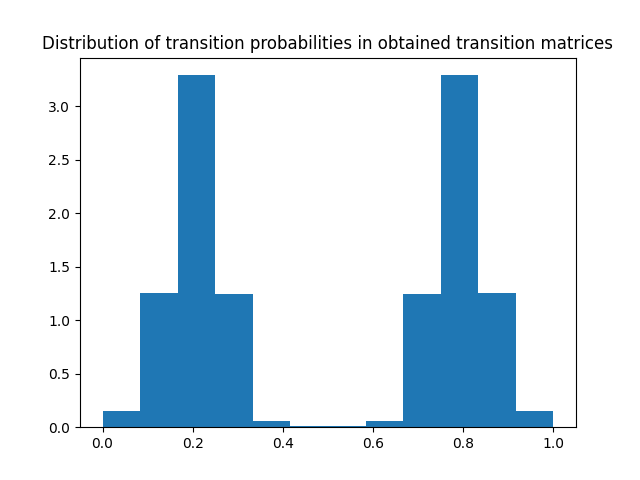}
\caption{Distribution of transition probabilities found in the obtained transition matrices across all 10 training instances. The transition matrices were obtained via a rank-2 NNCPD of the tensors. Since the target Bayesian network has a bit-flip probability of $p=0.2$ at each hidden node, the expectation is that the distribution is bimodal with peaks at $p=0.2$ and $p=0.8$.}
\label{fig:transition_probs}
\end{figure}
To verify the correctness of the obtained model and to illustrate how one might recover transition probabilities, we plot the distribution of transition probabilities in the transition matrices for all training instances. The transition matrices were obtained by computing an approximate rank-2 NNCPD of each intermediate tensor in the model via an MU-type method \cite{shashua2005}. Computing the NNCPD for both tensors along an edge produces one matrix each along the edge. The matrices are contracted together and the result is normalized appropriately so that it can be treated as a transition matrix.  The rank of the NNCPD for a tensor in the model corresponds to the number of hidden states at a hidden node in the associated hidden Markov model. The figure shows that the distribution of transition probabilities $p$ is strongly bimodal with modes at $p=0.8$ and $p=0.2$, which is in line with expectations given that the target Bayesian network passes its value downstream with probability $p=0.8$ and passes the complement of its value downstream with probability $p=0.2$. The variation in $p$ is due to the observation that an exact rank-2 NNCPD may not exist for a given degree-3 tensor. In the limit of a large number of data samples and an ideal optimization, we expect the tensors in the model to converge to the exact tensors, which admit an exact NNCP decomposition into the desired transition matrices.

\subsection{Hierarchical real-world data: phylogenetics}
Finally, we consider an example using real-world data in phylogenetics. In this numerical experiment, the objective is to construct a hierarchical model describing similarities between related organisms on a biomolecular level. The resulting network should provide a compact representation of the target distribution and a description of the clustering structure as well, which can be compared to the currently accepted classification. Using mitochondrial DNA (mtDNA) nucleotide sequence data from the cytochrome b (cyt b) gene for 16 different species in the taxonomic order Carnivora (obtained from the RefSeq project \cite{refseq}, see Tab.~\ref{tab:accession_codes}), we attempted to construct a phylogenetic tree corresponding to the data. The cytochrome b gene is 1140 base pairs (bp) for the species that were included in the dataset, and the gene is often used in phylogenetic studies because it offers good interspecies variation while remaining the same size for mammals \cite{castresana2001}. Species from the same taxonomic order were used so that the organisms are not too distant genetically, but distant enough to warrant multiple levels of clustering. Thus, this example attempts to hierarchically cluster the data.

\begin{table}
\caption{RefSeq accession codes used in the dataset.}
\label{tab:accession_codes}
\centering
\begin{tabular}{cc}
\toprule
Species & RefSeq Accession Code \\
\midrule
\textit{Canis lupus familiaris} & NC\_002008.4 \\
\textit{Panthera uncia} & NC\_010638.1 \\
\textit{Neofelis nebulosa} & NC\_008450.1 \\
\textit{Acinonyx jubatus} & NC\_005212.1 \\
\textit{Felis catus} & NC\_001700.1 \\
\textit{Phoca vitulina} & NC\_001325.1 \\
\textit{Ursus spelaeus} & NC\_011112.1 \\
\textit{Halichoerus grypus} & NC\_001602.1 \\
\textit{Arctocephalus forsteri} & NC\_004023.1 \\
\textit{Panthera tigris} & NC\_010642.1 \\
\textit{Ailurus fulgens} & NC\_011124.1 \\
\textit{Vulpes vulpes} & NC\_008434.1 \\
\textit{Mustela nivalis} & NC\_020639.1 \\
\textit{Ailuropoda melanoleuca} & NC\_009492.1 \\
\textit{Nandinia binotata} & NC\_024567.1 \\
\textit{Procyon lotor} & NC\_009126.1 \\
\bottomrule
\end{tabular}
\end{table}

We first translate the sequence nucleotides A, C, T and G into states from 0 to 3 and one-hot encode the data as four-dimensional vectors. The input sites in the initial network correspond to an organism and the input vectors are ordered by sequence position. In principle, if there is uncertainty in the sequence nucleotides, this uncertainty can be taken into account by using a probability vector instead of a one-hot vector. Since we have 16 species, the network has 16 input sites, and there are $|X|=1140$ samples corresponding to the aligned nucleotides in the gene. Most sequence positions are not phylogenetically significant and do not represent any grouping (for example, the invariant site positions), but in this example, we use all the data.

Here, we are more interested in obtaining a reasonable proposal for the hidden structure of the data as opposed to simply obtaining a ``good'' generative model in terms of the NLL. Phylogenetic trees represent hidden Markov models defined on a tree where the $4\times4$ transition matrices represent a nucleotide mutation probability. The training parameters are set so that the maximum bond dimension matches the input bond dimension, so $\chi_{max}=4$ and the resulting transition matrices have the desired size. Here, we must use an NMF-based method if we are interested in obtaining an interpretable model.

The maximum number of iterations in the training was set to $N=10000$ and the base learning rate was set to $\eta=0.0005$. For each scheme, 10 trials were conducted. Note that in this example, no assumptions were made on the underlying model: the tensors are not parametrized to simulate a specific model of evolution in contrast to what is usually done in phylogenetic studies with DNA substitution models like (in increasing order of complexity) the Jukes-Cantor model \cite{jukes1969}, the Hasegawa-Kishino-Yano model \cite{hasegawa1985}, and the generalized time-reversible model \cite{tavare1986}.
\begin{figure*}
\centering
\includegraphics[width=0.74\linewidth]{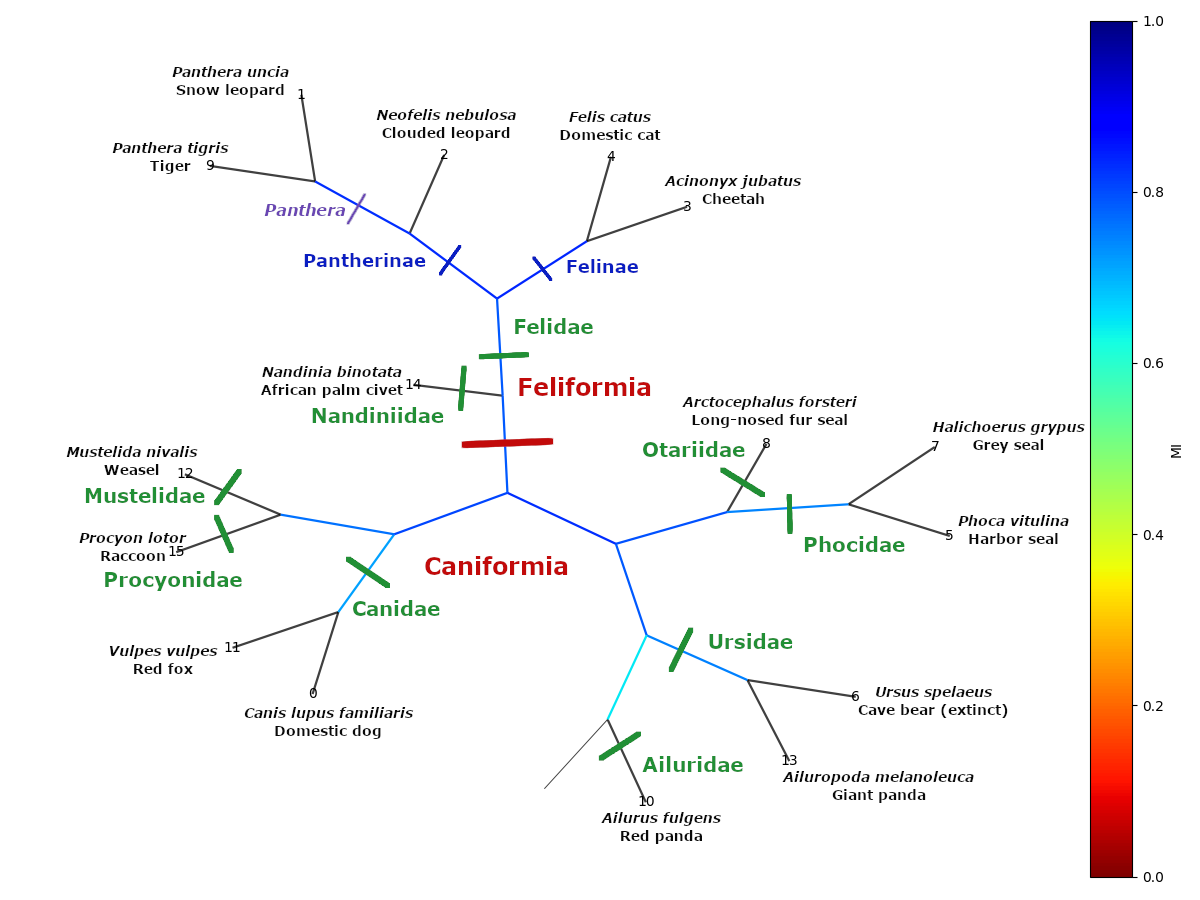}
\caption{Annotated tree structure for 16 species in order Carnivora, obtained using the hybrid scheme. At each terminal site, the species is identified. The colored divisions and labels denote taxonomic levels: red denotes suborders, green denotes families, blue denotes subfamilies, and purple denotes genera. Bond colors denote the normalized MI of each bond.}
\label{fig:carnivora_struct}
\end{figure*}

Fig.~\ref{fig:carnivora_struct} shows the best structure (in terms of NLL) obtained using the hybrid scheme, annotated with taxonomic distinctions. We found that the proposed network structure largely agrees with the existing literature describing the classification of members of order Carnivora \cite{mammal} -- one difference is that the position of the red panda should be closer to the weasels and raccoons than to the bears. Multiple levels of clustering are faithfully represented by the network structure: the feliform subtree is accurate down to the level of genus and correctly places the snow leopard in genus \textit{Panthera} \cite{johnson2006} (it has been historically classified in its own genus \textit{Uncia} \cite{mammal} and is labeled as such in the obtained dataset). However, since we do not consider any particular evolutionary model, the differences with existing literature concern the ordering of taxa in time (it is believed that dogs diverged first, then bears, which should place the weasels and raccoons closer to the seals \cite{flynn2005}). Furthermore, as we are limited to the use of sequence alignments from one gene (cytochrome b), we can extend the analysis by simply increasing the dataset size and considering additional sequence alignments in both mitochondrial and nuclear DNA.
\begin{figure}
\centering
\includegraphics[width=0.99\linewidth]{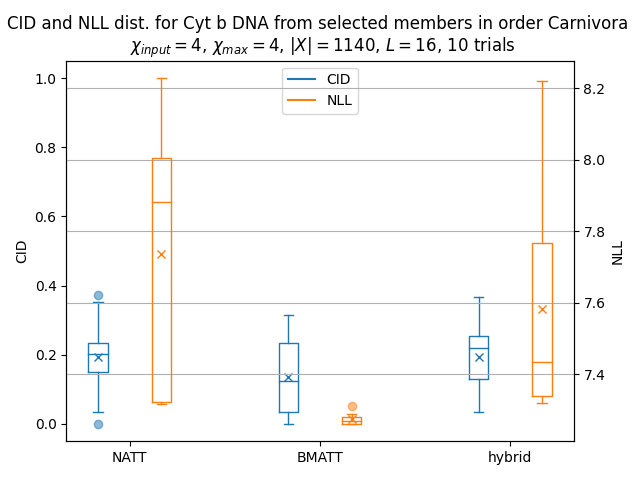}
\caption{Box plots of the cluster information distance (CID) (blue) and NLL (orange) for models trained on DNA sequence data. For each scheme, 10 trials were done. The NLL box plots consists of those 10 samples, while the CID box plots are taken across all pairs of trials, for a total of 45 (ways to choose 2 from 10) samples. Circles indicate outliers, crosses indicate means, and the horizontal line in the boxes denote the median.}
\label{fig:att_phylo_carnivora_tree_dist_cid_nll}
\end{figure}

Another aspect to consider would be whether or not the methods consistently produce solutions that are not too distant from each other. Since the network topology in Fig.~\ref{fig:carnivora_struct} is consistent with the literature, a low tree distance across trials would suggest that the method is indeed identifying the hidden structure of the data. In Fig.~\ref{fig:att_phylo_carnivora_tree_dist_cid_nll}, we find that this is the case, with the CID being consistently fairly low across all the schemes. While the BMATT scheme is also consistent in terms of NLL, the resulting scheme cannot be readily interpreted as a probability graph. Interestingly, even if the NLL appears to vary quite a bit for the NMF-based approaches, the obtained tree structures appear to be fairly close to each other. This further suggests that these methods prioritize obtaining a desirable structure.

\begin{figure}
\centering
\includegraphics[width=0.99\linewidth]{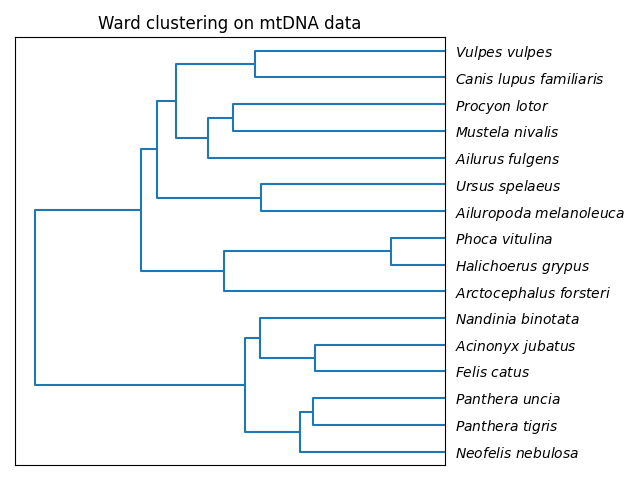}
\caption{Result of hierarchical clustering using the Ward method for the mtDNA data. Distances were computed using the Hamming metric on encoded nucleotides. The horizontal axis is a measure of cluster distance.}
\label{fig:ward_carnivora}
\end{figure}

In Fig.~\ref{fig:ward_carnivora}, we show the result of Ward clustering with the same mtDNA data. The Ward clustering also produces good agreement with the currently-accepted classification, and correctly positions the red panda closer to the musteloids than to the ursoids. However, it places the small cats (Felinae) further from the big cats/panthers, separating the members of the subfamily Felidae. Like with the ATT methods, there is no assumption of an underlying evolutionary model, so the Ward clustering does not reflect changes as a function of time. As mentioned previously, with the hierarchical clustering, we do not have access to a probabilistic model that describes the relationship between the inputs, which is possible with the tensor tree methods.

\begin{figure*}
\centering
\includegraphics[width=0.74\linewidth]{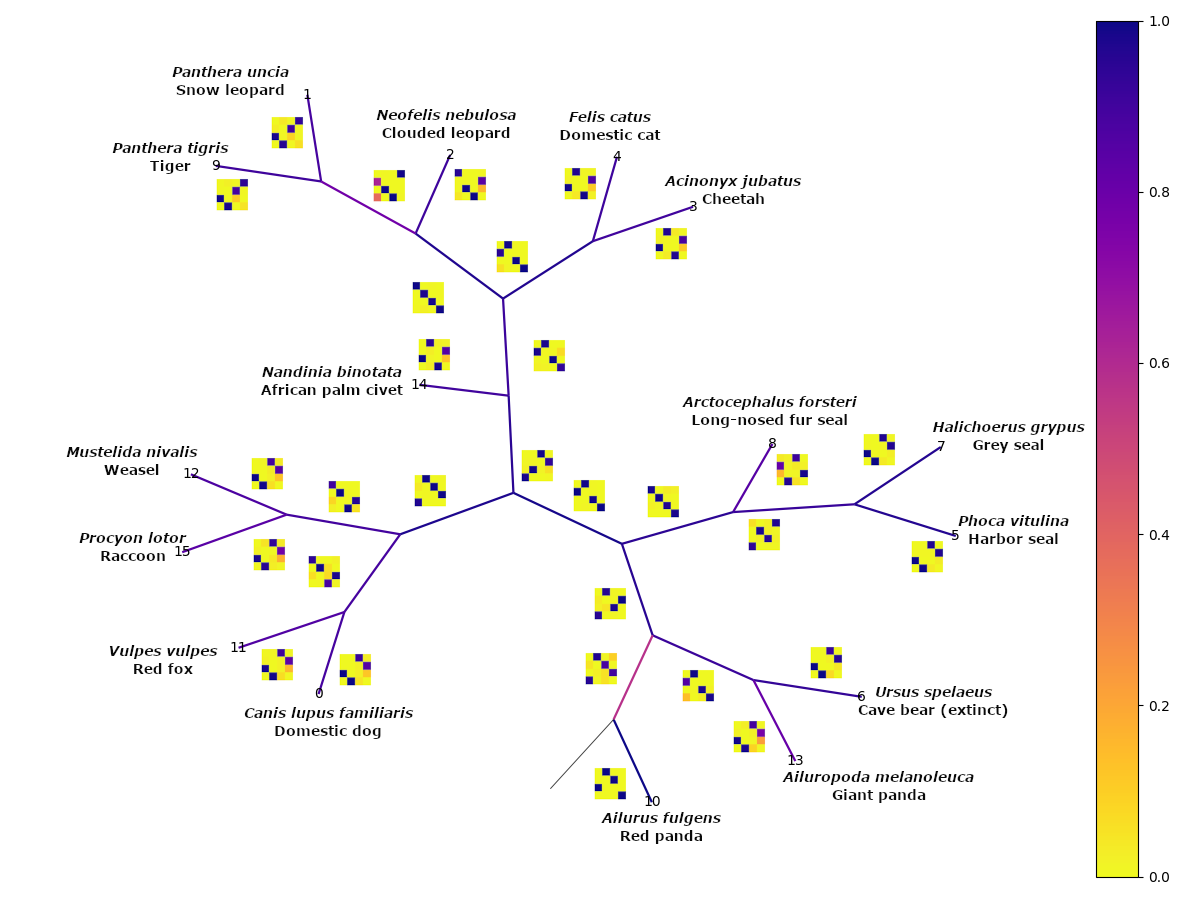}
\caption{Transition matrices obtained for the best tree structure for 16 species in order Carnivora, obtained using the hybrid scheme. At each terminal site, the corresponding species is identified. Along each bond, a heat map of the obtained transition matrix is provided that uses the same coloring rule as the bond colors, indicated by the bar in the right-hand side. Bond colors denote how close the transition matrix along the edge is to a permutation matrix, and this quantity is 1 if it is identical to a permutation matrix.}
\label{fig:carnivora_permutivity}
\end{figure*}
We can obtain an idea of the mutation probabilities between the sites in the network by computing the NNCPD of each tensor in the obtained NATT (shown in Fig.~\ref{fig:carnivora_struct}). Due to the ambiguity arising from relabeling of indices at each hidden site (equivalent to the insertion of permutation matrices along edges), we must consider how the obtained matrices are to be interpreted in a bit more detail.

It is generally assumed that mutation is not a common occurence, so it is expected that the transition matrices are described as having one element in each row and column that is close to one (i.e. that each matrix is close to a permutation matrix that accounts for the labeling ambiguity). Since the matrices are stochastic, the largest eigenvalue magnitude is 1, and this is reached for all eigenvalues when the transition matrix coincides with a permutation matrix. On the other hand, if the transition matrix is constant and all elements are identical, then there is one nonzero eigenvalue (which must be 1) and the remaining eigenvalues are zero. Thus, we can measure how close the matrix is to a permutation matrix by considering the smallest eigenvalue magnitude.

In Fig.~\ref{fig:carnivora_permutivity}, we observe that the bonds are all colored such that this measure is close to 1. This indicates that the matrices are all close to permutation matrices, and that the mutation probability is small. The heatmaps for each transition matrix also indicate this, and it is consistent with the expectation that the genetic distance between nearby organisms in the dataset is not significantly large.

\section{Summary and Discussion}
In order to obtain interpretable tensor trees for classical distributions, we replaced the double layer in the previous BMATT scheme \cite{harada2024} by a single layer with nonnegative tensors. The proposed NATT scheme uses nonnegative projected gradient descent and NMF to respect a nonnegative constraint on the variational parameters. We demonstrated that the new scheme performs almost as well as the double-layer scheme in minimizing the NLL and identifying the relational structures. By using the double-layer scheme to provide the single-layer scheme with an initial network structure, we can more consistently obtain good candidate structures than when considering only the single-layer scheme alone, while still keeping the network interpretable as a probabilistic graphical model. We also showed that the NATT and BMATT schemes are able to solve a variety of learning problems, including the XOR problem, which is characterized by an absence of two-point correlations in the input data, a random Bayesian network learning problem, where only a subset of sites are visible and the internal structure of the Bayesian network must be identified, and a real-world problem in phylogenetics, where mtDNA sequences are used to uncover a phylogenetic tree linking different members of order Carnivora. While the main advantage of the BMATT scheme centers around generative model quality (and by extension, sample quality) as measured by the NLL, the NATT scheme we propose in this work is useful when a probabilistic model is desirable: even if the BMATT scheme gives the correlation structure, it cannot provide an interpretable model.

From the various numerical experiments we have provided earlier, it is clear that the BMATT scheme performs best in terms of minimizing the NLL (see Fig.~\ref{fig:rand_l64}, Fig.~\ref{fig:bin_op_nll}, Fig.~\ref{fig:att_bayesian_rand_tree_dist_cid_nll}, and Fig.~\ref{fig:att_phylo_carnivora_tree_dist_cid_nll}). We attribute this behavior to two factors: first, the optimization problem for the BMATT is unconstrained, which is an indication of a generally easier problem, and second, the Born machine architecture has larger expressible power due to the absence of a nonnegativity constraint. This observation is possibly linked to why the BMATT scheme can reliably saturate the NLL bound when the bond dimension is large enough (as seen in Fig.~\ref{fig:rand_l64}).

However, the NATT scheme offers some unique advantages in comparison with the BMATT scheme. As the NATT scheme is based on the NMF, the strengths offered by the NMF in terms of interpretability and representation are inherited by the method: by construction, keeping the elements nonnegative allow us to interpret the NATT model more naturally. In most applications, we consider datasets generated by classical processes, so it is expected that a purely classical explanation for the data would be the most natural and ideal candidate model.

Finding the best structure to represent the data also provides clustering information in the process. Knowing the joint distribution $P(x)$ describing the data and being able to easily trace out variables (i.e. when any subgraph of the network representing $P(x)$ can be contracted efficiently, which is true when $P(x)$ is a tree) gives complete knowledge of the correlation structure of the data. For complicated data, we can leverage the BMATT scheme to find a good candidate initial structure and then use the NATT scheme to provide a compact representation of $P(x)$ that is entirely explainable classically. In this manner, we can benefit from the strengths of both schemes. We note that with datasets with a composition that changes over time, these plastic ATT approaches are able to adapt in accordance with the input data.

We conclude that this class of ATT-based methods is promising as a means to obtain the hidden structure of data. Further directions include probing the extent for which these ATT methods are useful, finding related methods that also learn temporal relationships and dependencies, and considering further applications that can drive knowledge discovery. As the proposed method relies on a constrained optimization problem, further improvements to the scheme, such as the choice of NMF method, also warrant further investigation. Another open problem, which may have some bearing on the classical-quantum distinction, is on the interpretation of EE-MI gaps and on the ease or difficulty of recovering an interpretable model from a Born machine representation.

\section{Data Availability Statement}
The data and code used to run the numerical experiments presented in this work can be found at the following URL: \href{https://github.com/redpinetree/att-ml}{https://github.com/redpinetree/att-ml}.

\begin{acknowledgments}
This work was partially supported by the joint project of Kyoto University and Toyota Motor Corporation, titled ``Advanced Mathematical Science for Mobility Society.'' K. A. would like to acknowledge the support of the Global Science Graduate Course (GSGC) program of the University of Tokyo. K. H. acknowledges the support from JSPS KAKENHI (Grant No. 20K03766 and 24K06886) and a Grant-in-Aid for Transformative Research Areas ``The Natural Laws of Extreme Universe—A New Paradigm for Spacetime and Matter from Quantum Information'' (KAKENHI Grants No. 21H05182 and No. 21H05191) from JSPS of Japan. T. O. acknowledges the support from JSPS KAKENHI (Grant No. 23H03818 and 22K18682), the Endowed Project for Quantum Software Research and Education, the University of Tokyo (https://qsw.phys.s.u-tokyo.ac.jp/), and The Center of Innovations for Sustainable Quantum AI (JST Grant No. JPMJPF2221). N. K. acknowledges the support from JSPS KAKENHI (Grant No. 23H01092). N.K. also appreciates useful comments from H. Shinaoka. Some of the computation in this work have been done using the facilities of the Supercomputer Center, the Institute for Solid State Physics, the University of Tokyo.
\end{acknowledgments}

\appendix
\section{Description of the cluster information distance}\label{app:cid}
To quantify the spread of proposed trees for a given scheme, we compute a tree distance measure between pairs of obtained networks and average over all the possible pairs. The tree distance measure we use is the cluster information distance (CID) \cite{smith2020}, which is a measure that generalizes the Robinsons-Foulds (RF) distance in a way that makes the tree distance more discriminative (i.e. takes more possible values, in this case, the CID is a real number instead of an integer like in the RF distance) and robust as a distance metric. This means that smaller changes between trees should correspond to smaller distances and larger changes should mean larger distances. The following description of the quantity closely follows \cite{smith2020}.

The CID is based on the clustering information associated with a tree topology. We consider a tree with leafset $X$ and a bipartition (``split'') $S=A|B$ on the tree. For a given split, we can define a clustering probability $P_{cl}(A)$ that a randomly-chosen leaf in $X$ is also in $A$, and this is just $P_{cl}(A)=|A|/|X|$ (and we can define a similar quantity for $B$). We only consider nontrivial splits that divide the tree into two nonempty partitions, so the number of splits possible in a tree corresponds to the number of internal edges. The entropy $H$ of a split $S$ is:
\begin{equation}
H(S)=-P_{cl}(A)\log P_{cl}(A)-P_{cl}(B)\log P_{cl}(B)
\end{equation}
We want to consider the distance between two trees $T_1$ and $T_2$ with the same leafset $|X|$. To do this, we must construct a matching $\mathcal{M}=\{(S_1,S_2)|S_2=f(S_1),f:\mathcal{S}_{T_1}\to\mathcal{S}_{T_2}\text{ is bijective},S_1\in\mathcal{S}_{T_1},S_2\in\mathcal{S}_{T_2}\}$ between $\mathcal{S}_{T_1}$, the set of splits of $T_1$ and $\mathcal{S}_{T_2}$, the set of splits of $T_2$. This effectively amounts to associating subtrees of $T_1$ and $T_2$ with each other. Then, if all subtrees are identical, then the trees are identical. The mutual information between partition $A_1$ in $T_1$ and $A_2$ in $T_2$ is:
\begin{equation}
I(A_1,A_2)=P_{cl}(A_1,A_2)\log\frac{P_{cl}(A_1,A_2)}{P_{cl}(A_1)P_{cl}(A_2)}
\end{equation}
Here, $P_{cl}(A_1,A_2)$ is the probability that a leaf belongs to $A_1$ in $T_1$ and $A_2$ in $T_2$, so $P_{cl}(A_1,A_2)=|A_1\cap A_2|/|X|$. Similar expressions can be defined for pairs $(B_1,B_2)$, $(A_1,B_2)$, and $(B_1,A_2)$. With these intermediate quantities, we can finally define a ``mutual clustering information'' score $I_{cl}(S_1,S_2)$ for two associated splits $S_1$ in $T_1$ and $S_2$ in $T_2$:
\begin{gather}
I_{cl}(S_1,S_2)=I(A_1,A_2)+I(B_1,B_2) \nonumber\\
+I(A_1,B_2)+I(B_1,A_2)
\end{gather}
To compute the CID, we must find the optimal matching $\mathcal{M}_{opt}$ between $T_1$ and $T_2$ that maximizes the sum of the scores $I_{cl}(S_1,S_2)$ over all paired splits:
\begin{equation}
I_{\Sigma,opt}=\max_{\mathcal{M}}\sum_{(S_1,S_2)\in\mathcal{M}}I_{cl}(S_1,S_2)
\end{equation}
This can be done efficiently by solving an assignment problem using the Hungarian algorithm, maximizing the total score. The total score $I_{\Sigma,opt}$ associated with the optimal matching is converted into a distance by subtracting it from an appropriate maximum value, which is half of the sum of the entropies of each split in $T_1$ and $T_2$. By rescaling against this maximum (as the minimum value of the CID is 0), we obtain a value normalized to be between 0 and 1.

\bibliography{main}% Produces the bibliography via BibTeX.

\end{document}